\documentclass[conference]{IEEEtran}
\IEEEoverridecommandlockouts

\usepackage{cite}
\usepackage{amsmath,amssymb,amsfonts}
\usepackage{graphicx}
\usepackage{booktabs}
\usepackage{textcomp}
\usepackage{xcolor}
\usepackage{url}

\begin{document}

\title{Physics-Informed Foresight Pruning for Sparse PINN Solvers of Nonlinear PDEs}

\author{\IEEEauthorblockN{Ahmad Ishaque Karimi}
\IEEEauthorblockA{\textit{School of Computing and}\\
\textit{Augmented Intelligence}\\
\textit{Arizona State University}\\
Tempe, USA \\
aikarimi@asu.edu}
\and
\IEEEauthorblockN{Uvini Balasuriya Mudiyanselage}
\IEEEauthorblockA{\textit{School of Computing and}\\
\textit{Augmented Intelligence}\\
\textit{Arizona State University}\\
Tempe, USA  \\
ubalasur@asu.edu}
\and
\IEEEauthorblockN{Kookjin Lee}
\IEEEauthorblockA{\textit{School of Computing and}\\
\textit{Augmented Intelligence}\\
\textit{Arizona State University}\\
Tempe, USA \\
Kookjin.Lee@asu.edu}

}

\maketitle

\begin{abstract}
Physics-informed neural networks (PINNs) often rely on over-parameterized models to optimize coupled solution and differential-residual objectives, leaving unclear how much capacity is necessary and what pruning should preserve. We study foresight pruning at initialization for sparse PirateNet PDE solvers. Standard neural tangent kernel spectrum-aware pruning (NTK-SAP) aims to preserve output-side training dynamics but may overlook parameters whose main influence arises through derivatives in the governing equations. We introduce physics-informed spectrum-aware pruning (PI-SAP), which assigns saliency using sensitivity of the PDE residual. Experiments on the Gray--Scott equations, complex Ginzburg--Landau equation, Burgers' equation, and linear convection equation show that PI-SAP more consistently preserves Gray--Scott residual fidelity and is competitive under aggressive sparsity. However, no criterion is uniformly optimal across equations or sparsity levels. Small-batch PINN-NTK diagnostics further show that residual fidelity, solution accuracy, and kernel conditioning are distinct objectives, motivating pruning methods that explicitly balance solution-side and residual-side training dynamics during optimization.
\end{abstract}


\section{Introduction}

Physics-informed neural networks (PINNs) approximate the solution of partial differential equations (PDEs) by combining data, initial or boundary conditions, and the governing residual in a differentiable training objective \cite{raissi2019}.
They are attractive for scientific machine learning because they avoid an explicit mesh at training time and represent the learned solution as a continuous function.
However, the same features that make PINNs flexible also make them difficult to train.
Deep coordinate networks can favor low-frequency modes \cite{rahaman2019}, derivative networks can be poorly conditioned at initialization, and the competing initial-condition and residual losses can create gradient imbalance \cite{wang2021,kim2021dpm}.
Recent convergence analyses also emphasize that residual minimization alone is not always sufficient to guarantee reliable approximation of the target physical solution \cite{doumeche2026}.

These difficulties are amplified in nonlinear PDEs with sharp gradients or coupled fields.
Reaction--diffusion and dispersive systems require a network to represent not only solution values but also the spatial and temporal derivatives appearing in the residual \cite{raissi2019,wang2021}.
Modern PINN architectures therefore use substantial model capacity to make optimization feasible; residual-adaptive PirateNets are one recent example \cite{wang2024piratenets}.
That capacity, however, increases memory and training cost. This computational burden is further amplified in parametric settings, where multi-query scenarios require resolving the PDE across a range of parameter configurations~\cite{cho2023hypernetwork,cho2024parameterized}. 
Foresight pruning \cite{lee2019snip,wang2020grasp,wang2023ntksap}, which removes weights at initialization before training begins, offers a way to reduce this redundancy while keeping the training protocol fixed.
The central question is therefore not simply whether a PINN can be pruned, but which notion of weight importance should be preserved.

Neural tangent kernel spectrum-aware pruning (NTK-SAP) provides a natural starting point for this question.
It is motivated by neural tangent kernel (NTK) theory, where the spectrum of the NTK controls gradient-descent training dynamics in wide networks \cite{jacot2018,wang2023ntksap}.
For ordinary supervised learning, preserving the output-side NTK spectrum is a principled way to preserve the dense model's training behavior under pruning.
For PINNs, however, output-side preservation may be incomplete.
Let $\theta$ denote the trainable network parameters.
A parameter with limited influence on the predicted fields $[u_\theta,v_\theta]$ can still strongly affect $[\partial_t u_\theta,\Delta u_\theta,\partial_t v_\theta,\Delta v_\theta]$ and therefore the PDE residual.
This observation motivates physics-informed spectrum-aware pruning (PI-SAP), a saliency rule that scores weights through the residual network rather than through the raw output map.

We evaluate this idea using residual-adaptive PirateNets \cite{wang2024piratenets}.
PirateNets initialize as shallow stable models and progressively open deeper nonlinear paths during training, making them a strong architecture for stiff PINN problems.
Our most complete experiment is the two-dimensional Gray--Scott reaction--diffusion system, whose stripe regime contains winding, derivative-sensitive interfaces.
We also report the complex Ginzburg--Landau equation, Burgers' equation, and convection equation experiments to test whether the observed behavior is specific to the Gray--Scott equations or reflects a broader cross-PDE pruning pattern.

The experiments reveal a qualified advantage rather than a universal winner.
PI-SAP improves residual fidelity of the Gray--Scott equations throughout the sparsity sweep and is particularly competitive at high sparsity, whereas NTK-SAP can be stronger at intermediate sparsity.
The two criteria therefore protect complementary aspects of training: output-side dynamics and residual-sensitive physics.

The PINN-NTK perspective formalizes this distinction because solution and residual losses evolve through different kernel blocks and can converge at different rates \cite{wang2022ntk}.
This motivates two diagnostic extensions.
Conditioning-aware NTK-SAP tests whether avoiding small eigenvalues and excessive spectral spread improves difficult modes, while PINN-block-aware SAP separately inspects small-batch solution and residual kernels.
Together, they test whether a useful sparse model must balance both dynamics rather than optimize either one in isolation.

The paper makes three contributions.
First, we benchmark NTK-SAP and PI-SAP across the Gray--Scott equations, complex Ginzburg--Landau equation, Burgers' equation, and convection equation PDEs to characterize when physics-informed saliency helps and when output-spectrum pruning remains competitive.
Second, we show that PI-SAP improves residual fidelity of the Gray--Scott equations at every pruning level and substantially reduces high-frequency error at 70\% pruning, supporting the claim that residual-aware saliency protects derivative-sensitive structure.
Third, through diagnostic experiments at 70\% pruning on the Gray--Scott equations, we show that improving NTK conditioning or preserving residual-side dynamics does not necessarily improve solution accuracy. This reveals a trade-off among solution accuracy, physics-residual fidelity, and optimization dynamics, motivating pruning criteria that explicitly balance these objectives.

\section{Technical Background}

\subsection{PINN Objective}

A PINN can use the same feed-forward architecture as a conventional supervised neural network: coordinates enter the network and predicted fields leave it.
The difference is the training signal.
A supervised network primarily compares its outputs with labeled targets, whereas a PINN also differentiates its outputs with respect to the input coordinates, substitutes those derivatives into the governing PDE, and minimizes the resulting residual together with initial or boundary errors.
Thus, the physics constraints supplement or replace much of the labeled supervision; they do not require a fundamentally different network architecture.

Let $z=(t,\mathbf{x})\in[0,T]\times\Omega$ denote a space--time coordinate and let $q_\theta:[0,T]\times\Omega\rightarrow\mathbb{R}^{d_q}$ be a neural solution parameterized by $\theta\in\mathbb{R}^{P}$.
Here $\mathbf{x}=x$ for a one-dimensional spatial domain and $\mathbf{x}=(x,y)$ for a two-dimensional domain; consequently, the implementation for the Gray--Scott equations uses $z=(t,x,y)$.
For a governing differential operator $\mathcal{N}$, define the pointwise physics residual as $r_\theta(z)=\mathcal{N}[q_\theta](z)$.
A typical PINN objective combines an initial/boundary constraint loss with an interior residual loss,
\begin{equation}
\mathcal{L}(\theta)=
\lambda_{\mathrm{ic}}\mathcal{L}_{\mathrm{ic}}(\theta)
+\lambda_{\mathrm{r}}\mathcal{L}_{\mathrm{r}}(\theta),
\end{equation}
where
\begin{equation}
\mathcal{L}_{\mathrm{r}}(\theta)
=\frac{1}{N_r}\sum_{i=1}^{N_r}
\left\|\mathcal{N}[q_\theta](t_i,\mathbf{x}_i)\right\|_2^2.
\end{equation}
Here $\lambda_{\mathrm{ic}},\lambda_{\mathrm{r}}\geq0$ are scalar loss weights, $\mathcal{L}_{\mathrm{ic}}$ enforces the prescribed initial and boundary data, and $\{(t_i,\mathbf{x}_i)\}_{i=1}^{N_r}$ are the $N_r$ interior collocation points.
For coupled systems such as the Gray--Scott equations or the complex Ginzburg--Landau equation, $q_\theta=[u_\theta,v_\theta]$ and $r_\theta=[r_u,r_v]$ contains one residual component for each field.
The residual depends on automatic differentiation through the network, so a pruning mask can change not only the predicted fields but also their derivatives.

\subsection{PirateNet Architecture and Pruning Scope}

We use a physics-informed residual adaptive network (PirateNet) as the
backbone of each PINN solver \cite{wang2024piratenets}. PirateNets are
designed to avoid unstable initialization of deep PDE residual networks
by beginning as shallow mappings and progressively introducing
nonlinear depth during training.

For the two-dimensional benchmarks, the input coordinate
$z=(t,x,y)$ is first transformed using fixed periodic encodings of
$x$ and $y$, followed by a trainable Fourier-feature embedding of
dimension 256. Two parallel dense transformations of this embedding
produce auxiliary feature streams $U$ and $V$. The embedded
representation then passes through three residual-adaptive blocks of
width 256 with Swish activations. For the one-dimensional benchmarks,
the coordinate input and activation function follow the corresponding
benchmark configuration; periodic encoding is used only when specified.

\begin{figure*}[t]
    \centering
    \includegraphics[width=\textwidth]
        {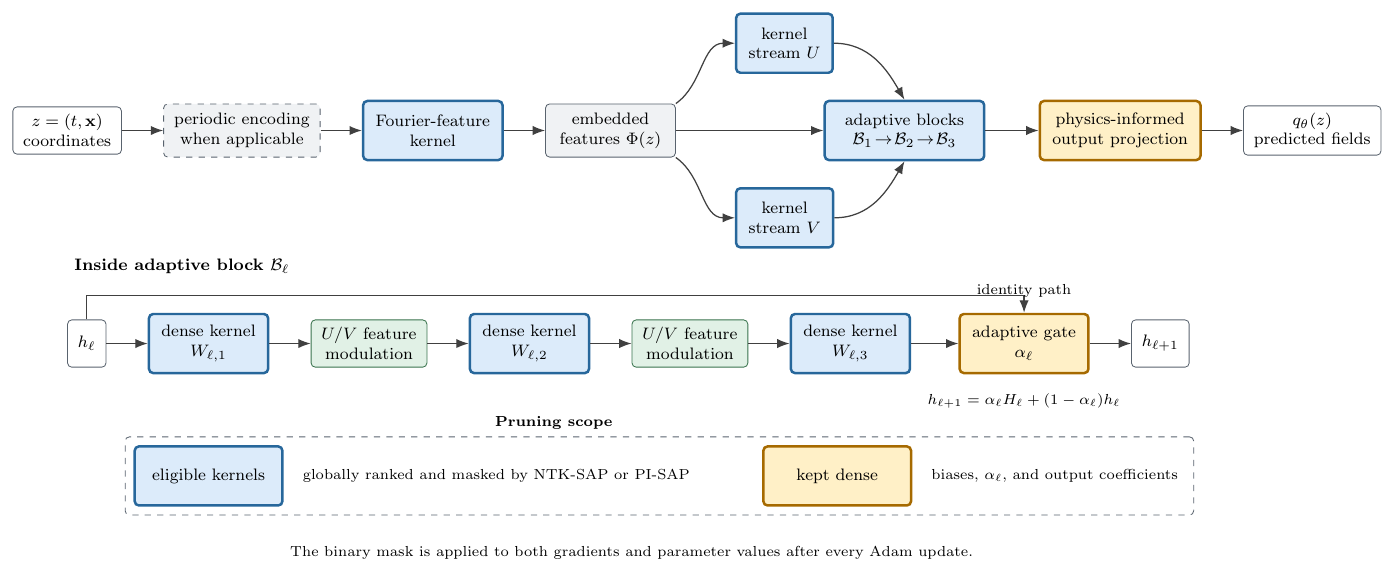}
    \caption{PirateNet backbone and pruning scope used in this study.
    The upper view summarizes the coordinate embedding, auxiliary
    feature streams, adaptive blocks, and output projection, while the
    lower view details one adaptive block. Blue components contain
    kernel parameters eligible for global pruning by NTK-SAP or PI-SAP.
    Adaptive gates $\alpha_{\ell}$, biases, and physics-informed output
    coefficients remain dense. The architecture follows the PirateNets
    design of Wang et al.~\cite{wang2024piratenets}; the pruning-scope
    representation is specific to our implementation.}
    \label{fig:piratenet-pruning-scope}
\end{figure*}

Figure~\ref{fig:piratenet-pruning-scope} summarizes the PirateNet
backbone and distinguishes parameters eligible for pruning from those
kept dense. Within each residual-adaptive block, three dense
transformations are interleaved with two modulation operations that
incorporate the auxiliary feature streams. If $h_{\ell}$ denotes the
input to block $\ell$ and $H_{\ell}$ denotes its nonlinear
transformation, the block output is

\begin{equation}
h_{\ell+1}
=
\alpha_{\ell}H_{\ell}
+
(1-\alpha_{\ell})h_{\ell},
\end{equation}

where $\alpha_{\ell}$ is a trainable scalar gate. Each gate is
initialized as $\alpha_{\ell}=0$, making every block an identity
mapping at initialization. As the gates evolve during optimization,
the nonlinear blocks are progressively introduced. A final linear
projection maps the learned representation to the predicted PDE
fields, such as
$q_{\theta}(z)=[u_{\theta}(z),v_{\theta}(z)]$
for the coupled Gray--Scott equations. Its coefficients are initialized
through a least-squares fit to the initial state of each temporal
window.

Pruning is applied globally only to trainable parameter leaves stored
as kernels. This eligible set includes the Fourier-feature kernel, the
two auxiliary-stream kernels, and the kernels within the three
residual-adaptive blocks. Because the dense layers use random weight
factorization, both factors in their kernel parameterization are
included in the eligible set. The reported pruning percentage is
computed globally over these eligible parameters. Biases, adaptive
gates $\alpha_{\ell}$, and the physics-informed output coefficients
are kept dense.

After saliency scoring, eligible parameters below the global threshold
are set to zero. The resulting binary mask is applied to both their
gradients and parameter values after every Adam update, preventing
pruned parameters from regrowing through optimizer momentum. NTK-SAP
and PI-SAP therefore use the same PirateNet architecture, eligible
parameter set, and mask-enforcement procedure; they differ only in the
quantity used to calculate parameter saliency.

\subsection{NTK View}

Let $f_\theta:\mathbb{R}^{d_{\mathrm{in}}}\rightarrow\mathbb{R}^{d_{\mathrm{out}}}$ be a differentiable feed-forward network; the PINN solution $q_\theta$ is one such map.
For an input batch $X$, let $f_\theta(X)$ denote its stacked output vector.
The Jacobian $J_\theta(X)=\partial f_\theta(X)/\partial\theta$ contains the derivative of every stacked output with respect to every trainable parameter.
The empirical neural tangent kernel (NTK) is
\begin{equation}
K(X,X)=J_\theta(X)J_\theta(X)^\top,\qquad
J_\theta(X)=\frac{\partial f_\theta(X)}{\partial\theta}.
\end{equation}
NTK theory connects the eigenspectrum of $K$ to gradient-descent training dynamics \cite{jacot2018}.
In the linearized NTK regime for squared loss, error components aligned with an eigenvector of $K$ decay at a rate controlled by the corresponding eigenvalue.
Small eigenvalues therefore create slow modes, while a wide eigenvalue spread creates uneven learning across modes.
In this paper, ``conditioning'' refers to this eigenspectral conditioning of the empirical kernel, for example through the condition number $\kappa(K)=\lambda_{\max}(K)/\lambda_{\min}(K)$, not to input normalization or an optimizer preconditioner.
This motivates NTK-preserving compression: if a sparse model preserves the relevant NTK spectrum, it should preserve important aspects of the dense model's optimization behavior.
NTK-SAP follows this principle by pruning connections that have little influence on an NTK-spectrum proxy \cite{wang2023ntksap}.
Related high-dimensional NTK compression theory also supports the idea that spectral equivalence can preserve convergence and generalization behavior under compression \cite{gu2024lossless}.

The PINN-NTK analysis of Wang et al. \cite{wang2022ntk} shows that PINN training can be understood through NTK blocks associated with different loss components, and that solution and residual terms can converge at different rates.
Thus, matching only the output-side dynamics is not guaranteed to preserve residual-side dynamics.
This motivates a pruning criterion that sees the differential operator.

\subsection{NTK-SAP Baseline}

NTK-SAP is a foresight pruning method: the sparse mask is selected before training.
The original method avoids forming a full NTK eigenspectrum by using a tractable trace/nuclear-norm proxy.
In our PINN implementation, the mask is computed at initialization and then enforced throughout training by masking both gradients and parameters, preventing pruned connections from regrowing through optimizer momentum.

At a high level, NTK-SAP introduces a mask $m\in[0,1]^P$ and scores each mask variable $m_j$ using the sensitivity of a perturbed output difference,
\begin{equation}
S_{\mathrm{NTK}}(m_j)=
\left|
\frac{\partial}{\partial m_j}
\left\|f_{\theta\odot m}(Z)-f_{(\theta+\Delta\theta)\odot m}(Z)\right\|_2^2
\right|,
\end{equation}
where $Z$ is a pruning input batch, $\Delta\theta$ is a small parameter perturbation, and $\odot$ denotes the elementwise (Hadamard) product.
Only ordinary kernel weights are eligible for pruning; biases, PirateNet $\alpha$ gates, and other initialization coefficients retain mask value one.
For target sparsity $s$, a global threshold is set at the $100s$-th percentile of all eligible scores: kernel weights at or below the threshold are assigned mask value zero, while higher-scoring weights are retained.
Thus, the rule removes the eligible weights judged least influential to the output-dynamics proxy.

\section{Proposed Method}

\subsection{PI-SAP}

For a coupled PDE residual $r_\theta=[r_u,r_v]$ evaluated on a residual collocation batch $X_r$, PI-SAP scores
\begin{equation}
S_{\mathrm{PI}}(m_j)=
\left|
\frac{\partial}{\partial m_j}
\left\|r_{\theta\odot m}(X_r)-r_{(\theta+\Delta\theta)\odot m}(X_r)\right\|_2^2
\right|.
\end{equation}
PI-SAP uses the same global ranking and target sparsity $s$ as NTK-SAP.
The practical intent is simple: protect the connections that most affect satisfaction of the governing PDE.
For reaction--diffusion systems, this means preserving weights that influence the derivative-sensitive diffusion and reaction terms, not merely weights that influence the raw output magnitude.

\subsection{Diagnostic Extensions}

Beyond residual saliency, we consider two diagnostic objectives that use small-batch empirical kernels during mask selection.
Their definitions are independent of the governing PDE and can be applied whenever differentiable solution and residual maps are available.
The first, conditioning-aware NTK-SAP, examines the empirical kernel of a selected network output map $f_\theta$.
Instead of only preserving a dense-model spectral proxy, it favors masks whose small-batch empirical NTK has larger average scale and less spectral collapse.
This is not an architectural preconditioner and does not form the full training-set NTK.
For a small pruning batch $Z$, it computes $J=\partial f_\theta(Z)/\partial\theta$ and forms
\begin{equation}
K=JJ^\top
\end{equation}
as a compact diagnostic/objective.
For eigenvalues $\lambda_i^\epsilon=\max(\lambda_i,\epsilon)$, the implemented objective has the form
\begin{equation}
\phi(K)=
\log \bar{\lambda}^\epsilon
-\alpha_c(\log \lambda_{\max}^\epsilon-\log \lambda_{\min}^\epsilon)
-\alpha_s\,\mathrm{std}(\log \lambda_i^\epsilon),
\end{equation}
where $\epsilon>0$ is an eigenvalue floor, $\bar{\lambda}^\epsilon$ is the mean clipped eigenvalue, and $\alpha_c,\alpha_s\geq0$ weight the condition-number and log-spectrum-spread penalties, respectively.
The mask saliency is computed from $|\theta_j\,\partial\phi/\partial\theta_j|$.

The second, PINN-block-aware SAP, separates solution-side and residual-side empirical kernels.
For the shared small batch $X$ used by the current implementation, define $J_{\mathrm{sol}}=\partial q_\theta(X)/\partial\theta$ and $J_{\mathrm{res}}=\partial r_\theta(X)/\partial\theta$; for a two-field system these stack the component Jacobians $[J_u;J_v]$ and $[J_{r_u};J_{r_v}]$, respectively.
The corresponding kernel blocks are
\begin{align}
K_{\mathrm{sol}} &= J_{\mathrm{sol}}J_{\mathrm{sol}}^\top,\\
K_{\mathrm{res}} &= J_{\mathrm{res}}J_{\mathrm{res}}^\top.
\end{align}
These are practical small-batch analogs of the $K_{uu}$ and $K_{rr}$ blocks used in the PINN-NTK view.
The block-aware objective combines $\phi(K_{\mathrm{sol}})$ and $\phi(K_{\mathrm{res}})$ and penalizes mismatch in their mean log-eigenvalue scales.
Cross terms $K_{ur}=J_{\mathrm{sol}}J_{\mathrm{res}}^\top$ and $K_{ru}=K_{ur}^\top$ are left for future work.

\section{Experimental Setup}
The main experiments on the Gray--Scott equations use the scaled system
\begin{align}
u_t &= \epsilon_1 \Delta u + b_1(1-u) - c_1uv^2,\\
v_t &= \epsilon_2 \Delta v - b_2v + c_2uv^2.
\end{align}
The domain is $\Omega=[-1,1]^2$ with $t\in[0,2]$, 101 saved snapshots, and a $200\times200$ reference grid.
The parameters are $\epsilon_1=0.2$, $\epsilon_2=0.1$, $b_1=40$, $b_2=100$, and $c_1=c_2=1000$.
With the code scaling, this corresponds approximately to the standard feed/kill values for the Gray–Scott equations, $F=0.04$ and $k=0.06$ because $b_1=1000F$ and $b_2=1000(F+k)$.

All models for the Gray--Scott equations use PirateNet with 3 residual adaptive blocks, width 256, Swish activation, Fourier features of dimension 256, Adam optimization, GradNorm weighting, and causal time marching over 10 temporal windows.
The completed sparsity sweep for the Gray--Scott equations uses five pruning levels: 10\%, 30\%, 50\%, 70\%, and 90\%.
The diagnostic extension runs use the same stripe-like regime at 70\% pruning with a shorter budget of 60,000 steps per window.

For the complex Ginzburg--Landau equation, we report dense, NTK-SAP, and PI-SAP runs across the same five pruning levels.
For Burgers' equation and convection equation, we use prior cross-PDE studies averaged over seeds 0--4 with 50k Adam steps.
These latter experiments are not as deep as the Gray--Scott study, but they are useful for testing whether the observed sparsity trends are equation-specific.

We report relative $L^2$ error for each field and the mean solution error $(e_u+e_v)/2$ when two fields are present.
For the Gray--Scott equations and the Ginzburg--Landau equation, we also report mean PDE residual $(\ell_{r_u}+\ell_{r_v})/2$.
Runtime is reported as wall-clock training time for the current masked implementation.
Because the implementation masks dense arrays rather than invoking sparse kernels, pruning is not expected to produce proportional wall-clock speedups.

\section{Results}

\subsection{Gray--Scott Equations: Main Case Study}

Table~\ref{tab:gray-scott} is the main result.
PI-SAP has lower mean solution error at 10\%, 30\%, 70\%, and 90\% pruning, while NTK-SAP is slightly better at 50\%.
The larger distinction appears in the physics residual: PI-SAP has lower mean PDE residual at every pruning level.

\begin{table}[t]
\caption{Results for the Gray--Scott equations in the stripe-forming regime. $\Delta$ denotes the PI-SAP mean-error change relative to NTK-SAP; negative values indicate improvement.}
\label{tab:gray-scott}
\centering
{\scriptsize
\setlength{\tabcolsep}{2.5pt}
\begin{tabular}{lccccc}
\toprule
Prune & NTK err. & PI err. & $\Delta$ & NTK res. & PI res. \\
\midrule
10\% & $1.1439{\times}10^{-2}$ & $7.5343{\times}10^{-3}$ & -34.1\% & $1.2996{\times}10^{-5}$ & $1.1403{\times}10^{-5}$ \\
30\% & $1.7896{\times}10^{-2}$ & $1.7555{\times}10^{-2}$ & -1.9\% & $1.5345{\times}10^{-5}$ & $1.5200{\times}10^{-5}$ \\
50\% & $4.3067{\times}10^{-2}$ & $4.4727{\times}10^{-2}$ & +3.9\% & $2.6953{\times}10^{-5}$ & $2.1316{\times}10^{-5}$ \\
70\% & $2.7291{\times}10^{-1}$ & $1.9430{\times}10^{-1}$ & -28.8\% & $5.1482{\times}10^{-5}$ & $4.8413{\times}10^{-5}$ \\
90\% & $4.3493{\times}10^{-1}$ & $3.7866{\times}10^{-1}$ & -12.9\% & $3.8921{\times}10^{-4}$ & $3.8713{\times}10^{-4}$ \\
\bottomrule
\end{tabular}
}
\end{table}

At 10\% pruning, PI-SAP reduces mean error by about 34.1\% relative to NTK-SAP.
At 70\%, it reduces mean error by about 28.8\%.
At 90\%, PI-SAP still improves mean error by about 12.9\%, even though both methods have entered a degraded high-sparsity regime.
The 50\% exception is important: residual-aware saliency is not automatically better for every sparsity.
However, the high-sparsity behavior supports the main hypothesis that physics-informed saliency becomes more valuable when the sparse model has fewer redundant paths.

The residual trend is more consistent.
PI-SAP improves the mean PDE residual at every pruning level, with the largest residual reductions at 10\% and 50\%.
This does not mean lower residual always implies lower solution error, but it does show that residual-sensitive pruning better preserves sampled physics consistency across the full sparsity sweep for the Gray--Scott equations.
Both methods exhibit a sharp jump in solution error between 50\% and 70\%, suggesting that the practical accuracy threshold for this PirateNet configuration lies in that interval.
Runtime stays in the 16.4--18.0 hour range for these runs of the Gray--Scott equations, with no systematic wall-clock advantage from masking.

\subsection{High-Frequency Diagnostic for the Gray--Scott Equations}

The stripe regime of the Gray--Scott equations is visually governed by sharp spatial interfaces.
To check whether the L2 trends reflect loss of high-frequency content, we computed terminal-time Fourier errors at 70\% pruning from saved checkpoints.
The high-frequency mask keeps spatial modes whose radial frequency is at least 0.35 of the maximum discrete frequency.

\begin{table}[t]
\caption{Terminal high-frequency errors for the Gray--Scott equations at 70\% pruning.}
\label{tab:hf}
\centering
{\small
\begin{tabular}{lcc}
\toprule
Method & $u$ high-freq. & $v$ high-freq. \\
\midrule
Dense PirateNet & 0.0227 & 0.0277 \\
NTK-SAP 70\% & 0.7238 & 0.7749 \\
PI-SAP 70\% & 0.3591 & 0.4132 \\
\bottomrule
\end{tabular}
}
\end{table}

PI-SAP roughly halves the terminal high-frequency error relative to NTK-SAP for both species.
This strengthens the interpretation that residual-aware saliency better protects the derivative-sensitive connections needed to resolve stripe interfaces.

\subsection{NTK-Block Diagnostics for the Gray--Scott Equations}

Table~\ref{tab:extensions} summarizes the diagnostic extensions for the Gray--Scott equations at 70\% pruning.
These runs use a shorter 600k-step budget and should not replace the completed sweep in Table~\ref{tab:gray-scott}.

\begin{table}[t]
\caption{Diagnostic extensions for the Gray--Scott equations at 70\% pruning and 600k total steps.}
\label{tab:extensions}
\centering
{\scriptsize
\setlength{\tabcolsep}{4pt}
\begin{tabular}{lcccc}
\toprule
Method & $u$ err. & $v$ err. & $r_u$ loss & $r_v$ loss \\
\midrule
Dense & 0.0264 & 0.0489 & $2.262{\times}10^{-4}$ & $8.219{\times}10^{-5}$ \\
NTK-SAP & 0.1786 & 0.3229 & $1.973{\times}10^{-4}$ & $6.971{\times}10^{-5}$ \\
Cond. NTK-SAP & 0.1966 & 0.3516 & $2.589{\times}10^{-4}$ & $9.279{\times}10^{-5}$ \\
PI-SAP & 0.2006 & 0.3578 & $6.667{\times}10^{-4}$ & $2.086{\times}10^{-4}$ \\
PINN-block SAP & 0.2249 & 0.4021 & $1.605{\times}10^{-4}$ & $5.671{\times}10^{-5}$ \\
\bottomrule
\end{tabular}
}
\end{table}

The diagnostic result is intentionally nuanced.
Under this shorter budget, original NTK-SAP has the best solution accuracy among pruned models.
PINN-block-aware SAP has the lowest residual losses among pruned models but the worst solution error.
The conditioning-aware NTK-SAP variant also does not improve over the original NTK-SAP result in this setting.
Thus, these extensions are best interpreted as diagnostic probes: block-level conditioning terms can improve one target, such as sampled residual loss, without preserving the full solution field.
This is the central reason for treating sparse PINN pruning as a multi-objective design problem rather than selecting a single saliency score.

\subsection{Complex Ginzburg--Landau Equation}

Table~\ref{tab:gl} reports the Ginzburg--Landau comparison, including the dense baseline.
The result is not a duplicate of Gray--Scott.
Both pruning methods improve over dense at 10\%, NTK-SAP is better at 30\% and 50\%, and PI-SAP is better at 70\% and 90\%.

\begin{table}[t]
\caption{Ginzburg--Landau summary. Dense mean error is $2.8017{\times}10^{-2}$ and dense residual is $1.4557{\times}10^{-5}$.}
\label{tab:gl}
\centering
{\scriptsize
\setlength{\tabcolsep}{2.5pt}
\begin{tabular}{lccccc}
\toprule
Prune & NTK err. & PI err. & Better & NTK res. & PI res. \\
\midrule
10\% & $2.6062{\times}10^{-2}$ & $2.5384{\times}10^{-2}$ & PI & $1.0831{\times}10^{-5}$ & $1.1401{\times}10^{-5}$ \\
30\% & $2.8769{\times}10^{-2}$ & $3.4257{\times}10^{-2}$ & NTK & $1.4901{\times}10^{-5}$ & $1.4704{\times}10^{-5}$ \\
50\% & $4.6634{\times}10^{-2}$ & $5.4244{\times}10^{-2}$ & NTK & $2.7359{\times}10^{-5}$ & $2.6487{\times}10^{-5}$ \\
70\% & $6.5615{\times}10^{-2}$ & $5.8078{\times}10^{-2}$ & PI & $7.9350{\times}10^{-5}$ & $8.2189{\times}10^{-5}$ \\
90\% & $1.8836{\times}10^{-1}$ & $1.7434{\times}10^{-1}$ & PI & $4.7924{\times}10^{-3}$ & $3.7924{\times}10^{-3}$ \\
\bottomrule
\end{tabular}
}
\end{table}

The Ginzburg--Landau results support a more careful claim than ``PI-SAP always wins.''
PI-SAP improves mean error by about 11.5\% relative to NTK-SAP at 70\% pruning and by about 7.4\% at 90\%, but NTK-SAP is better at 30\% and 50\%.
This agrees with the trend observed for the Gray--Scott equations that residual-aware saliency is most valuable when pruning pressure is aggressive, while also showing that the balance between output and residual saliency depends on the PDE and sparsity level.
The runtime remains near 10 hours for all runs, again indicating that mask-based pruning does not automatically translate to wall-clock acceleration without sparse kernels.

\subsection{Burgers' Equation and Convection Equation}

The Burgers' equation and convection equation studies provide additional cross-PDE context.
They were run over seeds 0--4 for 50k Adam steps.
Table~\ref{tab:burgers} reports the best pruned Burgers' equation result for each network width, while Table~\ref{tab:convection} reports the best pruned convection equation result for each $\beta$--width setting.
In both tables, we report only the best pruned result rather than every sparsity level to keep the comparison compact.

\begin{table}[t]
\caption{Best pruned results for Burgers' equation across network widths.}
\label{tab:burgers}
\centering
{\scriptsize
\setlength{\tabcolsep}{3.5pt}
\begin{tabular}{lccc}
\toprule
Width & Dense & Best pruned & Method \\
\midrule
128 & $1.0849{\times}10^{-2}$ & $3.7532{\times}10^{-3}$ & PI 30\% \\
256 & $3.1485{\times}10^{-2}$ & $5.4160{\times}10^{-3}$ & PI 30\% \\
512 & $3.0051{\times}10^{-1}$ & $2.5324{\times}10^{-3}$ & NTK 90\% \\
\bottomrule
\end{tabular}
}
\end{table}

\begin{table}[t]
\caption{Best pruned results for the convection equation across transport parameter $\beta$ and network width.}
\label{tab:convection}
\centering
{\scriptsize
\setlength{\tabcolsep}{3.5pt}
\begin{tabular}{ccccc}
\toprule
$\beta$ & Width & Dense & Best pruned & Method \\
\midrule
1  & 128 & $1.2495{\times}10^{-2}$ & $5.2080{\times}10^{-3}$ & PI 30\% \\
   & 256 & $1.1271{\times}10^{-2}$ & $5.0310{\times}10^{-3}$ & PI 70\% \\
   & 512 & $1.3225{\times}10^{-1}$ & $7.2160{\times}10^{-3}$ & NTK 90\% \\
\midrule
5  & 128 & $1.6948{\times}10^{-2}$ & $6.8240{\times}10^{-3}$ & NTK 50\% \\
   & 256 & $9.2240{\times}10^{-3}$ & $9.4590{\times}10^{-3}$ & PI 30\% \\
   & 512 & $9.6970{\times}10^{-3}$ & $7.4770{\times}10^{-3}$ & PI 70\% \\
\midrule
10 & 128 & $2.4645{\times}10^{-2}$ & $1.8089{\times}10^{-2}$ & PI 50\% \\
   & 256 & $1.8334{\times}10^{-2}$ & $1.0124{\times}10^{-2}$ & NTK 10\% \\
   & 512 & $3.6643{\times}10^{-2}$ & $1.7202{\times}10^{-2}$ & PI 30\% \\
\bottomrule
\end{tabular}
}
\end{table}

These supporting results show two patterns.
First, pruning can act as structural regularization: the best pruned models for Burgers' equation outperform the dense baselines at all three widths, and the effect is especially strong at width 512.
Second, the best saliency rule is not fixed.
PI-SAP is best for Burgers' equation at widths 128 and 256, while NTK-SAP is best at width 512.
For the convection equation, the preferred method also varies with both $\beta$ and network width: PI-SAP is best in several settings, while NTK-SAP is selected in others, such as $\beta=1$ at width 512, $\beta=5$ at width 128, and $\beta=10$ at width 256.

As an additional high-$\beta$ stress test at width 128, we also evaluate $\beta=15$ and $\beta=20$.
In both cases, the best pruned models still improve over the dense baselines: NTK-SAP at 70\% sparsity reduces the mean relative $L^2$ error from $5.3298{\times}10^{-2}$ to $2.5010{\times}10^{-2}$ for $\beta=15$, and NTK-SAP at 50\% sparsity reduces it from $1.3302{\times}10^{-1}$ to $7.9886{\times}10^{-2}$ for $\beta=20$.
Overall, these results reinforce the main paper message: physics-aware pruning is useful, but sparse PINN selection should be evaluated across equation type, physical regime, sparsity, width, and training stiffness.

\section{Discussion}

Across the four PDE families, the results support a consistent but nontrivial conclusion.
Pruning is viable for PINN solvers, but the useful pruning criterion depends on what part of the physics-constrained training dynamics is under pressure.
NTK-SAP is a strong output-dynamics baseline because it is tied to spectral preservation.
PI-SAP becomes attractive when residual-sensitive structures matter, especially in high-sparsity settings for the Gray--Scott equations and the complex Ginzburg--Landau equation.
The results for Burgers' equation and the convection equation show that this is not a universal dominance claim: the best method can shift with width, advection strength, and sparsity.

The most important observation for the Gray--Scott equations is the separation between solution error, residual error, and high-frequency error.
PI-SAP lowers the mean PDE residual across the complete sparsity sweep for the Gray--Scott equations and reduces high-frequency error at 70\%.
However, the 70\% diagnostic extensions show that optimizing residual-side or kernel-conditioning criteria alone is not sufficient.
A sparse PINN can achieve a lower residual loss on sampled collocation points while having worse relative solution error.
This is consistent with the PINN-NTK view that different loss blocks can evolve at different rates \cite{wang2022ntk}.

For scientific machine learning, this suggests a practical evaluation standard.
Sparse PINN solvers should be judged by at least three quantities: solution error, physics residual, and a task-aware spectral or morphological diagnostic.
For reaction--diffusion systems, high-frequency spatial error is an effective diagnostic because pattern fidelity is lost first at the interfaces.
For advection-dominated systems, stability across seeds and aggressive sparsity may be more informative.
This multi-metric view is also aligned with the workshop goal of efficient and trustworthy AI for scientific applications: efficiency without physical fidelity is not enough, and residual reduction without solution fidelity can be misleading.

\section{Limitations and Reproducibility}

The experiments should be interpreted as evidence for a pruning principle, not as a final sparse-kernel acceleration result.
The implementation enforces masks on dense JAX arrays, so the reported wall-clock times measure training stability and overhead under fixed software infrastructure rather than hardware-level sparse speedup.
This is why the runtimes for the Gray--Scott equations and the complex Ginzburg--Landau equation remain nearly constant across sparsity levels.
Actual acceleration would require sparse matrix kernels, structured sparsity, or compiler support that exploits the mask pattern during training and inference.
For the present study, this design choice is intentional: it isolates the numerical effect of the pruning criterion from the engineering effect of a sparse backend.

The second limitation is statistical.
The studies on Burgers' equation and the convection equation are averaged over five seeds, but the results for the Gray--Scott equations and the complex Ginzburg--Landau equation are treated primarily as completed benchmark runs rather than large multi-seed sweeps.
The Gray--Scott equations are expensive because each run uses 10 causal windows and long training budgets, and the purpose of the study is to compare pruning behavior under a fixed, reproducible solver setup.
Future work should repeat the high-sparsity benchmark settings for the Gray--Scott equations and the complex Ginzburg--Landau equation across seeds, since the most interesting differences occur near the accuracy-degradation threshold where initialization effects may matter.

The third limitation concerns the PINN-block diagnostics.
The small-batch kernels $K_{\mathrm{sol}}$ and $K_{\mathrm{res}}$ are practical approximations, not full training-set NTKs.
This is necessary because explicitly constructing full NTK blocks for every collocation point and every residual component would be computationally prohibitive.
The diagnostic coefficients were not extensively tuned, so these runs should be read as a first stress test of the objective design rather than a final optimized method.
However, the diagnostic result is still useful: it shows that improving the residual-block diagnostic can lower sampled residual loss while worsening solution error.
The next pruning objective should therefore condition both blocks jointly, potentially including cross terms $K_{ur}$ and $K_{ru}$, rather than optimizing either output or residual dynamics in isolation.

\section{Conclusion}

This paper studies physics-informed foresight pruning for sparse PINN solvers across nonlinear PDEs, with the Gray--Scott equations as the most complete benchmark and the complex Ginzburg--Landau equation, Burgers' equation, and the convection equation as cross-PDE validation.
The results show that sparse PirateNet solvers can retain useful accuracy under substantial pruning, and that residual-aware PI-SAP improves physics residuals for the Gray--Scott equations across all tested sparsities while providing the largest solution benefits in high-sparsity regimes.
Cross-PDE experiments confirm that the advantage is not uniform: NTK-SAP remains competitive and sometimes superior, particularly at intermediate sparsity or in specific width/parameter settings.
The NTK-block diagnostics further show why future pruning objectives should balance solution-side and residual-side dynamics instead of optimizing either one in isolation.
The resulting direction is a conditioning- and physics-aware pruning framework for scientific ML models that reduces parameter count while remaining interpretable through NTK diagnostics and validated by PDE-specific fidelity metrics.

\end{document}